\documentclass{article}

\usepackage{arxiv}

\usepackage[utf8]{inputenc} 
\usepackage[T1]{fontenc}    
\usepackage{hyperref}       
\usepackage{url}            
\usepackage{booktabs}       
\usepackage{amsfonts}       
\usepackage{nicefrac}       
\usepackage{microtype}      
\usepackage{lipsum}
\usepackage{graphicx}

\title{A Methodological Review of Visual Road Recognition Procedures for Autonomous Driving Applications}

\author{
  Kai Li~Lim\\
  The REV Project\\
  The University of Western Australia\\
  Perth, Australia\\
  \texttt{kaili.lim@uwa.edu.au} \\
   \And
 Thomas~Bräunl\\
  The REV Project\\
  The University of Western Australia\\
  Perth, Australia\\
  \texttt{thomas.braunl@uwa.edu.au} \\
}

\begin{document}
\maketitle

\begin{abstract}
The current research interest in autonomous driving is growing at a rapid pace, attracting great investments from both the academic and corporate sectors. In order for vehicles to be fully autonomous, it is imperative that the driver assistance system is adapt in road and lane keeping. In this paper, we present a methodological review of techniques with a focus on visual road detection and recognition. We adopt a pragmatic outlook in presenting this review, whereby the procedures of road recognition is emphasised with respect to its practical implementations. The contribution of this review hence covers the topic in two parts -- the first part describes the methodological approach to conventional road detection, which covers the algorithms and approaches involved to classify and segregate roads from non-road regions; and the other part focuses on recent state-of-the-art machine learning techniques that are applied to visual road recognition, with an emphasis on methods that incorporate convolutional neural networks and semantic segmentation. A subsequent overview of recent implementations in the commercial sector is also presented, along with some recent research works pertaining to road detections.
\end{abstract}


\section{Introduction}\label{secintro}
The field of autonomous driving is attracting much attention lately, ever since its feasibility was established in the 2007 DARPA Urban Challenge \cite{RN330}. These days, technological and automotive corporates are expediting the announcements of autonomous vehicles to the consumer market alongside electric vehicles, which are also becoming imminently available. From a research standpoint, this area is also well-documented in the literature. Conventional systems often rely on radar and subsequently, LiDAR to detect road kerbs and edges, but with the advancement of computer vision, cameras are quickly replacing these sensors as the preferred sensor to detect and recognise roads. Cameras also benefit from being versatile and low-cost, in addition to its ubiquity which enables the deployment of visual autonomous driving on a larger scale. 

Using cameras for road detection and recognition however introduces challenges whereby it is heavily reliant on the robustness of the image processing algorithms to accurately recognise road regions, often in real-time. This is unlike LiDAR and/or radar-based approaches that usually relies on the processing of the sensor's measurement values to classify roads from non-road regions around the vehicle. A robust algorithm for road recognition should account for the dynamic variations of road types, conditions and illumination, as well as seasonal and weather changes pertaining to the road scene, while being able to accurately perform road classification.  

A comprehensive background study and review on this topic was presented in 2013 \cite{RN167}, whereby the authors have discussed the problems faced by visual road detection and recognition, particularly in the two categories of common roads --- structured (with lane markings) and unstructured (without lane markings). Like most visual computing problems, visual road detection is also susceptible to variations of lighting, along with weather and seasonal changes. Solutions to visual road detection are similar to most visual computing approaches, whereby a captured image frame will first undergo preprocessing where to reduce noise and other imperfections. The image will subsequently have its features detected and extracted using algorithms such as Scale-Invariant Feature Transform (SIFT) \cite{RN279} or Speeded-Up Robust Features (SURF) \cite{RN278}, and then distinguishing these features as road areas. These extracted features will correspond to areas on the image where road and non-road regions are distinguished. This is essentially a sequential process of image preprocessing, feature extraction and model fitting \cite{RN166}. Aspects that are unique to visual road detection are the prevalence of the horizon and the vanishing point of the road. The horizon is the boundary of the frame where the sky and land meets; and the vanishing point is the point where the road converges to. These aspects along with the road establish the region of interest (ROI) where visual processing can be carried out. 

With the advent of deep learning, convolutional neural networks (CNN) are increasingly being incorporated into road detection algorithms to train and classify roads with improved accuracy \cite{RN169,RN170,RN187}. Using CNN introduces high processing requirements and increases the complexity of the algorithm. Such an algorithm is usually trained offline at a dedicated server or workstation to obtain a dataset related to the driving environment. Datasets for road detection includes KITTI \cite{RN171} and Daimler \cite{RN172}. Conversely, there are proposals of fast and simple algorithms that perform road recognition without the need of another computer, and does not require training \cite{RN173,RN156}. This further classifies visual road recognition into supervised and unsupervised algorithms, indicating the presence or absence thereof a training classifier within the algorithm. This being said, the application of CNNs onto an image processing problem means that computation using the graphics processing unit (GPU) is rapidly gaining in popularity. The high parallelism of GPU architectures is especially suitable for the parallel nature of visual and deep learning applications such as road detection. The Nvidia DRIVE \cite{RN174} solution is a testament to this, whereby an industrial GPU maker is currently developing GPU solutions to autonomous driving that is centred around deep learning and computer vision. Additionally, other specialised hardware such as Mobileye's Automated Driver Assistance System (ADAS) is the core technology utilised by many of their 27 automotive manufacturer partners across 313 models for their autonomous driving feature \cite{RN221}.

This paper is organised as follows. Section \ref{secalgo} presents works covering the procedural implementation of road detection using standard classification methods, including the detection of vanishing points, region of interests, image classification and model fitting. Methods that incorporates machine learning, particularly on CNN and its methods, are presented in Section \ref{seclearning}. Section \ref{seccommercial} presents commercial implementations with regards to products and courses with road recognition, and Section \ref{secrecent} presents the current trends and works in recent years for road detection before this the concluding remark is presented in Section \ref{secconclusion}.  

\section{Conventional Methods} \label{secalgo}
Road recognition for autonomous driving generally follows the methods described in sections \ref{sechorizon} to \ref{secmodel} in a chronological order. More specifically, many implementations start with a preprocessing stage to filter image noise and other inconsistencies, followed by a horizon detection algorithm to crop the horizon so that the image is processed only at the road portion that is below the horizon line. Researchers may then use vanishing point detection to orient or localise the vehicle on the road. Regions of interests may be used to isolate the process on road or lane marking edges for image classification. During image classification, methods include using a combination of either edge detection, colour histograms, textural comparison, machine learning or neural networks; which is then typically classified in binary (non-road/road regions) through methods such as Gaussian filtering or confidence voting. Finally, the navigational boundaries are marked on the edges with lines and subsequently plotting the centre path for the vehicle to drive on.

\subsection{Horizon Detection} \label{sechorizon}
\begin{figure*}[ht]
	\centering
	\includegraphics[width=0.7\linewidth]{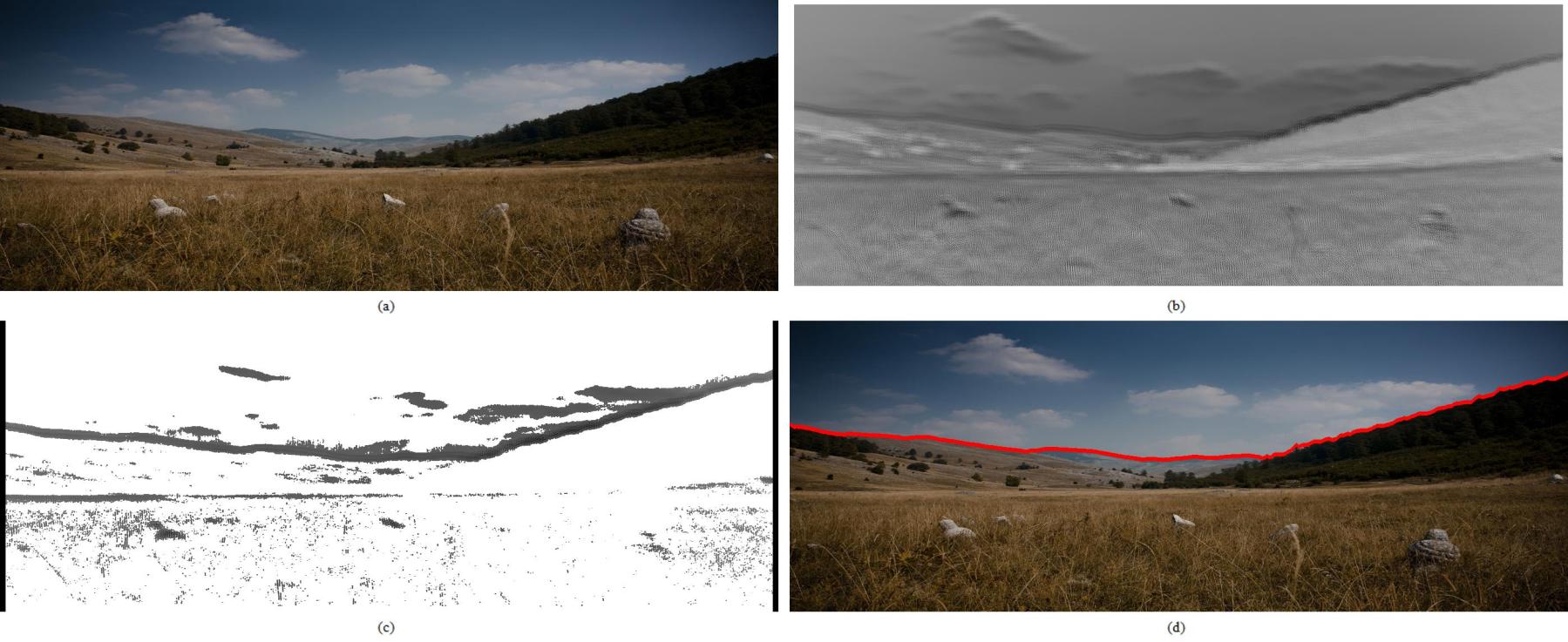}
	\caption{\label{fig:edgeless}An edge-less horizon detection algorithm generates a dense classifier score image (DCSI) (b) from a query image (a) using trained classifiers, where a threshold is then applied (c) before it plots the horizon line (d). Reprinted with permission from \cite{RN160}}
\end{figure*}

Horizon detection algorithms are typically used for road detection to crop the image frame, thereby reducing overall computational requirements. Horizon detection is generally applied onto a preprocessed image to calculate the boundary of the skyline for an image frame. The area above the horizon will be isolated and ignored for the rest of the computation. A fast horizon detection algorithm is favourable to present minimal processing and memory footprint with reference to the overall visual road recognition solution. While the fastest and simplest approach may be to fix the horizon at a constant pixel location according to the camera's orientation, this assumes that the vehicle is always traversing on a perfectly even terrain with no variations of pitch nor incline, which is generally unachievable under normal driving circumstances. Horizon detection is either edge-less or edge-based \cite{RN157}. Edge-less approaches use edge classification whereby the horizon is detected by removing non-horizon edges through the refinement of the edge map; in an edge-less approach, each pixel location is classified according to their probability of it being on the horizon. An edge-based approach was proposed by Lie et al. \cite{RN176}, where a multi-stage graph was generated from an edge map using a dynamic programming algorithm. An edge-less approach was proposed by Ahmad et al. \cite{RN160} which instead uses a classification map to find the horizon line that incorporates machine learning and dynamic programming. According to the authors, using an edge-less approach will not require the assumption of the horizon line being close to the top of the image frame. An example of the process of using an edge-less algorithm is illustrated in Figure \ref{fig:edgeless}. Ahmad et al. went on to propose a method that fuses edge-based and edge-less approaches \cite{RN157}, outperforming both edge-based and edge-less approaches. On the deep learning front, Verbikas and Whitehead \cite{RN178} incorporated CNN into horizon detection, which outperforms other classifiers in their experiments in accuracy. CNN was applied to train classifiers to recognise sky and ground features with spatial feature extractors.

\subsection{Vanishing Point Detection} \label{secvanishing}
\begin{figure*}[ht]
	\centering
	\includegraphics[width=\linewidth]{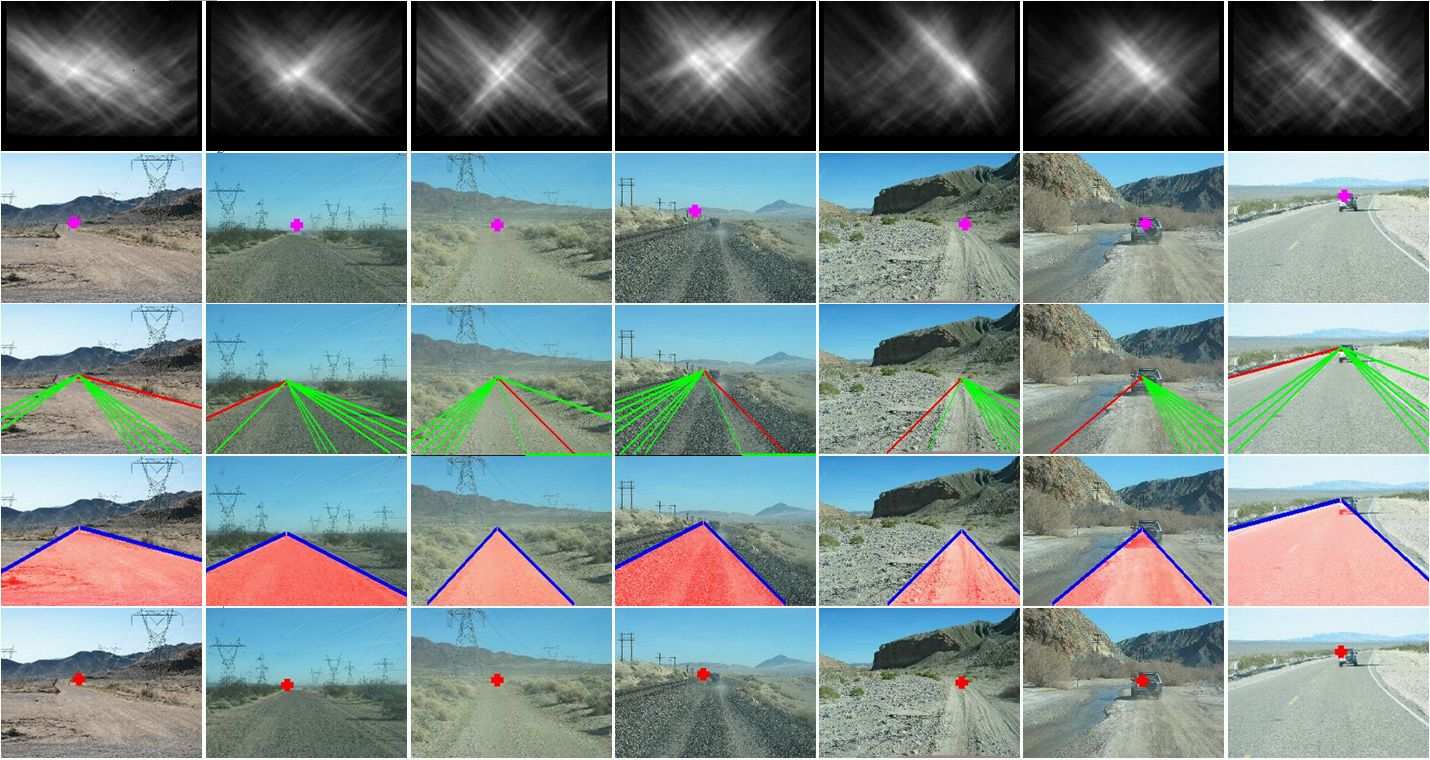}
	\caption{\label{fig:vanishingpoint}Vanishing point estimation of seven desert road images showing the LASV algorithm. Reprinted with permission from \cite{RN181}.}
\end{figure*}

This is typically used in tandem or as an alternative to horizon detection to localise the vehicle with respect to the image frame. The vanishing point is a point on an image where a pair or more parallel lines in 3D space converges to. According to Rother \cite{RN179}, the detection of vanishing point consists of an accumulation step and a search step. The accumulation step clusters line segments that share a common vanishing point, and the search step searches for dominant line clusters. Rother noted that the random smaple consensus (RANSAC) method could be used to speed up vanishing point detection, to which Bazin and Pollefeys \cite{RN180}  proposed an approach that uses only three lines to achieve this. They proposed this approach for a three degrees of freedom (3DoF) robotic manoeuvrability system similar to a ground vehicle. This effectively enables the system to estimate its rotation based on its captured visual lines and vanishing point. Kong, Audibert and Ponce \cite{RN181} described an approach to road detection that centres around vanishing point detection, and their method is as illustrated in Figure \ref{fig:vanishingpoint}. They proposed the Locally Adaptive Self-Voting (LASV) algorithm that estimates the vanishing point based on a confidence model in a local region for texture orientation estimation. While it is more accurate than conventional voting approaches, Zhu et al. \cite{RN183} noted that this approach does not perform well in suburban environments with dense roadside vegetation and fixtures. They subsequently utilised a colour histogram method to compare the captured image against an \textit{a priori} model to obtain the vanishing point. Still, line voting remains a popular method to achieve vanishing point detection and it is also used by Zhu et al. and other recent works \cite{RN184,RN185,RN332,RN182,RN189}. 

\subsection{Region of Interest Isolation} \label{secregion}
ROI isolation methods are used as a popular approach to recognise road segments from non-road segments. An ROI is usually identified and defined in frames before these image segments are classified. Instead of needing to process the entire frame, using an ROI isolates image processing to a frame's specific region to further reduce computing requirements. This may be used in tandem with horizon detection where certain regions below the horizon line are designated as the ROI. For road recognition, the ROI is generally a definitive region that encompasses both road and non-road regions or lane markings, as classification can then be drawn from processing that ROI. While conventional ROIs are usually fixed at a predetermined location on a frame, this assumes that the road boundary will always be on the same frame location \cite{RN193,RN192}. To circumvent this, adaptive ROI algorithms were proposed as a more robust solution that adjusts to illumination changes \cite{RN163} or the location of vanishing points \cite{RN188}. 

\subsection{Image Classification} \label{secimage}
Image classification is used in this context to classify an image into road and non-road areas using a binary classifier \cite{RN156}. Roads are recognised through a combination of the road lane markings and road boundary. Lane markers are usually painted in high contrast from the road surface to be conspicuous to the drivers, and visual processing also benefits from this whereby good edge detection results can be obtained more easily. Conversely, variations in road lane appearances such as colours, lines and condition from wear and tear may pose a challenge for lane detection algorithms \cite{RN164}. Works that attempt to circumvent these variations include \cite{RN195,RN196}. Some roads, especially non-gravel roads, are unstructured and have no lane markings at all. In these circumstances, lane detection algorithms will not work, and road boundary detection algorithms will be applied. On urban roads, the road boundary is perceived as the region where the asphalt meets an unpaved ground. Works that detect road boundaries may also use kerbs \cite{RN197} or highway barriers \cite{RN198}. More robust road boundary detection algorithms aim to work across different road types, including dirt roads and snow roads \cite{RN181}; or variations to illumination and weather such as night driving and rain \cite{RN199,RN200}. Techniques used for both road lane and boundary detection may vary, but they may also share some similarities, especially with the usage of edge detection algorithms. For instance, stereoscopic sensors can be used to perceive the tangible road boundaries in urban areas, which works as an alternative to radar or LiDAR \cite{RN201}, or in conjunction for added robustness \cite{RN202}. Non-urban roads commonly share the same plane as non-road areas, so such detection algorithm should be purely appearance-based. An example of such an algorithm was proposed by Crist\'{o}foris et al. \cite{RN156}, where they applied a mixture of Gaussians (MOG) model onto an image's ROI on HSV colour space that is converted from RGB. This more commonly known as the Gaussian mixture model (GMM). The GMM is a form a Bayesian classification, which performs decision making using the probability theory based on the maximum likelihood estimation (MLE). As its name suggests, a GMM a combination of several Gaussian distributions and hence the MLE is derived from the weight sum of the Gaussians as the probability density function \cite{RN325}. In the context of road detection, GMM analyses the colour distribution of the road and estimates the colour model based its similarities with similar model groups. Gaussian models are widely used as a supervised learning approach to image classification.   Other works that incorporate a Gaussian model include \cite{RN203,RN204}. Alkhorshid et al. \cite{RN205} used a histogram obtained from the calculation of frequency distribution of pixel values, and subsequently used candidate training to classify whether or not the ROI is fully, partially or not part of a road region; this is modelled after the AdaBoost classifier \cite{RN207} that is employed to minimise weighted errors. Other methods of classification include using textural features \cite{RN208,RN210}, which assumes a homogeneous road texture that is compared to non-road textures. 

\subsection{Model Fitting} \label{secmodel}
\begin{figure}[ht]
	\centering
	\includegraphics[width=0.5\linewidth]{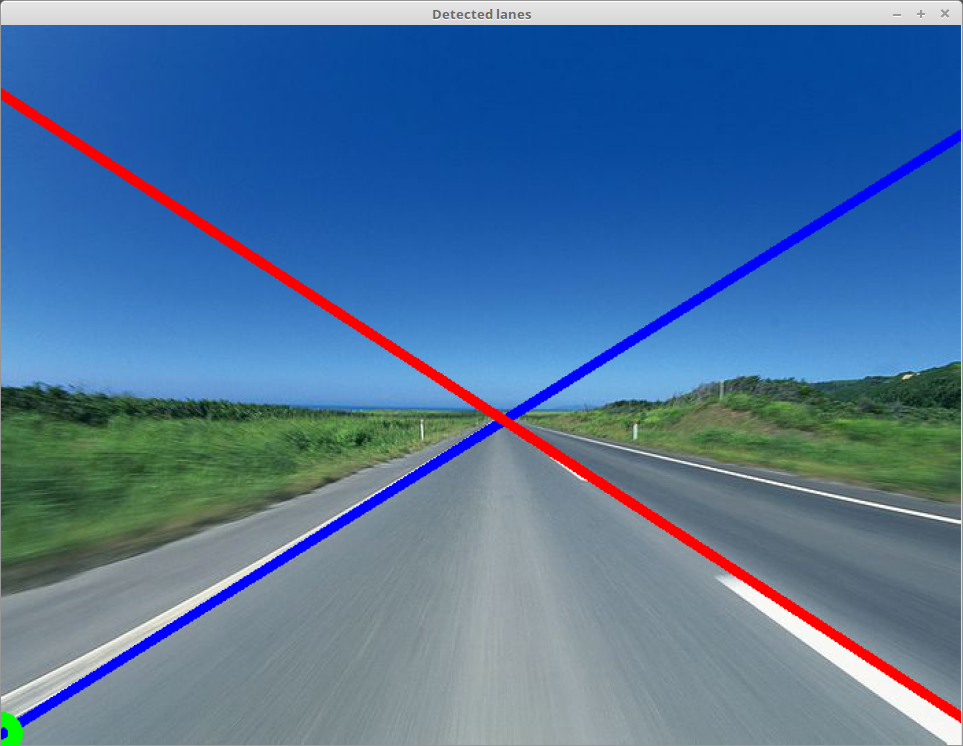}
	\caption{\label{fig:lanedetect}Lane detection with Hough transform performed using OpenCV}
\end{figure}

Once road and non-road regions have been classified in the image, navigational boundaries must be marked (fitted) to prevent the vehicle from veering off course. These boundaries can be marked more easily on marked roads using edge detection algorithms that benefit from the large gradient values. Edge filters such as the Sobel and Canny filters are commonly used \cite{RN211,RN212,RN213}. Regions of the high gradient can then be plotted according to the filters' results. These plots will result in the road or lane boundary, and they can either be parametric, semi-parametric or non-parametric \cite{RN166}. Parametric models comprise mostly of straight lines \cite{RN181}; semi-parametric models comprise of splines \cite{RN214} and polynomial curves \cite{RN215,RN216}; and non-parametric models comprise of continuous arbitrary lines \cite{RN156}. Urban roads typically have well-defined lane markings and boundaries, hence these navigational boundaries can be marked with a parametric or semi parametric model, effectively reducing computation complexity. Rural and unpaved roads may require the use of semi or non-polynomial models. Outliers are commonly present with fitting models, therefore it is also common to implement RANSAC for outlier rejection at this stage. Aly \cite{RN203} used RANSAC to fit lines and splines in his lane detection algorithm, and it is applied after performing a simplified Hough transform for lane line counting. With the road/lane boundaries marked, the vehicle can then be guided to drive at the centre of the road/lane by finding the distance between its left and right boundaries. In addition to road boundaries, model fitting can also be applied for marking the horizon and path for navigation, such as the approach used by Crist\'{o}foris et al. \cite{RN156}. A popular technique of model fitting in road recognition is the Hough transform \cite{RN222}. The Hough transform is a shape analysis technique typically used to extract shape features from an image. Road detection and recognition works commonly apply Hough line transform on an edge-detected image to detect lane markings and road edges according to the aforementioned line models. This achieves road segmentation, splitting road sections for vehicles to recognise areas such as lanes and non-road areas. The Hough transform uses a voting procedure to fit lines. This means that each point that may correspond to a line section votes for the likelihood that a line section may be from. More votes are cast when more points lie on the same line, and lines with higher votes will be fitted \cite{RN223}. Figure \ref{fig:lanedetect} illustrates an example of lane detection using Hough transform, the blue and red lines mark the left and right edge of the lane respectively, where a driving rule can be established to ensure that the vehicle does not cross these boundaries while driving. Works that employ the Hough transform or a similar voting approach to classify road and non-road regions are \cite{RN181,RN187,RN183}.

\section{Learning Methods} \label{seclearning}
Many works are implementing learning methods these days for road recognition, where support vector machines (SVMs) \cite{RN281,RN283}, neural networks \cite{RN282} and AdaBoost \cite{RN284} are among the more commonly implemented approaches. A neural network approach was proposed as early as 2003 by Conrad and Foedisch \cite{RN280} using Matlab's Neural Network Toolbox. The authors compared this approach to a SVM approach and noted that while SVMs are more accurate, their computation times are long for road classification tasks. It is noted that SVM has historically been a mainstream learning method for road recognition, with newer implementations incorporating dynamic programming to account for the changes in road scenes \cite{RN285}. Learning methods on road recognition is seeing a rise in popularity also due to the increased demand in autonomous driving, and the KITTI benchmark suite also includes are road and lane detection evaluation benchmark since 2013 \cite{RN286}. This benchmark categorises road scenes according to a combination of roads types including urban marked, unmarked, multiple marked lanes/roads. There are currently 336 benchmark submissions for the categories to date, using various methods such as SVM \cite{RN290}, CNN \cite{RN288,RN289} and may incorporate other sensors such as LiDAR \cite{RN287}.

Due to the increased availability of parallel computers nowadays, recent works are more commonly implementing convolutional neural networks for road recognition in favour of SVM and custom networks. CNNs are feedforward neural networks with more than one convolution layer. This convolution layer applies a function repeatedly over the output of other functions, which greatly benefit from parallel computation. For road recognition, CNNs are commonly used for object detection and image segmentation, which are used on road scenes to detect and discern areas in the image that encompasses an object, where in addition to roads, it also segments other elements such as pedestrians, vehicles, vegetation and road signs. This process is commonly known as semantic segmentation. Semantic segmentation classes objects in an image according to its pixels, thereby improving the system's comprehension towards road scenes in addition to road classification over conventional learning methods, allowing for a more holistic autonomous driving system that also incorporates features such as road sign understanding, pedestrian and vehicle detection, and collision avoidance \cite{RN292,RN291,RN272}. Thoma \cite{RN293} and Garcia-Garcia et al. \cite{RN297} published surveys of semantic segmentation, which include a good background study of the underlying approaches of semantic segmentation. While the KITTI benchmark suite has not yet incorporated a semantic segmentation benchmark, it does provide a list of resources of KITTI images with semantic labels. MultiNet \cite{RN291} is an example that combines semantic segmentation, classification and detection for road scenes that is consolidated from the same encoder to minimise time redundancy. Results were tested and benchmarked on KITTI's road dataset where it is found to be capable of real-time processing. 

Semantic segmentation uses neural networks and is hence trained and tested on datasets. The review paper by Garcia-garcia et al. \cite{RN297} provides a detailed analysis of the datasets used in semantic segmentation, categorising them into 2D, 2.5D and 3D datasets. The KITTI \cite{RN294}, CamVid \cite{RN326} and Cityscapes \cite{RN296} datasets are more commonly associated with training and testing semantic segmentation for urban road scenes. The KITTI Vision Benchmark Suite is an actively maintained project by the Karlsruhe Institute of Technology and the Toyota Technological Institute of Chicago. Images were obtained by driving a car around Karlsruhe, Germany, covering a variety of road scenes. This benchmark suite spans across several categories including optical flow, stereo vision, visual odometry, and road/lane detection. The road/lane detection evaluation benchmark consists of 289 and 290 training and test images respectively. Additionally, the CamVid dataset was proposed in 2008 containing 367 training and 233 testing images. Images were obtained from the City of Cambridge, England by driving a car around. This dataset was created for semantic segmentation and each training image pixel is labelled with a different shade of grey corresponding to one of the twelve classes for road scene objects, forming the ground truth. Finally, the Cityscapes dataset was obtained from 50 cities in Germany from a road vehicle across three seasons. This is a 5000-image dataset whereby 2975, 500 and 1525 images are categorised for training, validation and testing respectively. The authors of Cityscapes made comparisons to KITTI and CamVid and noted that semantic labelling can be easily achieved with their smaller datasets, and that the Cityscapes dataset provides a better challenge for new semantic segmentation approaches with its much larger dataset. Table \ref{tabdatasets} summarises the datasets presented here according to its publication data and number of sample images for training and testing.

\begin{table}[ht]
	\renewcommand{\arraystretch}{1.3}
	\caption{Summary of road datasets presented}
	\label{tabdatasets}
	\centering
	\begin{tabular}{c c c c}
		\hline
		& & \multicolumn{2}{c}{\bfseries Image Samples} \\
		\bfseries Dataset & \bfseries Year & \bfseries Training & \bfseries Testing\\
		\hline
		KITTI-Road \cite{RN294}		& 2013  		& 289 		& 290 \\
		CamVid \cite{RN326}			& 2009 			& 367 		& 233 \\
		Cityscapes \cite{RN296}		& 2016 			& 2975 		& 1525 \\
		\hline
	\end{tabular}
\end{table}

\begin{figure}[ht]
	\centering
	\includegraphics[width=4.7in]{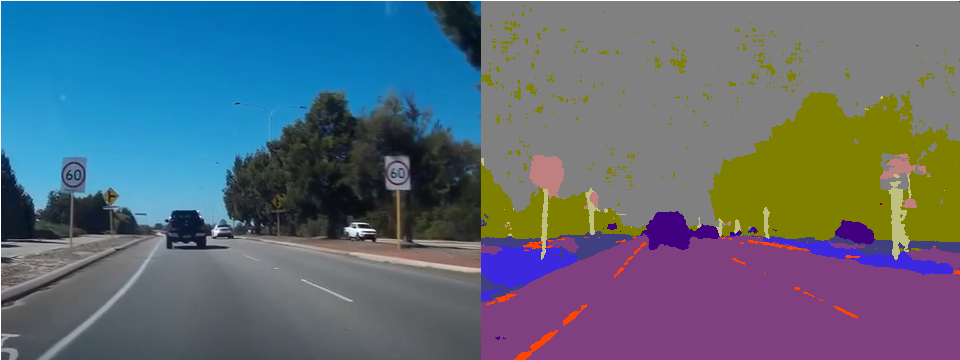}
	\caption{SegNet's input (left) and output (right) for a typical Western Australian road scene.}
	\label{figsegnet}
\end{figure}

\begin{figure*}[!ht]
	\centering
	\includegraphics[width=4.7in]{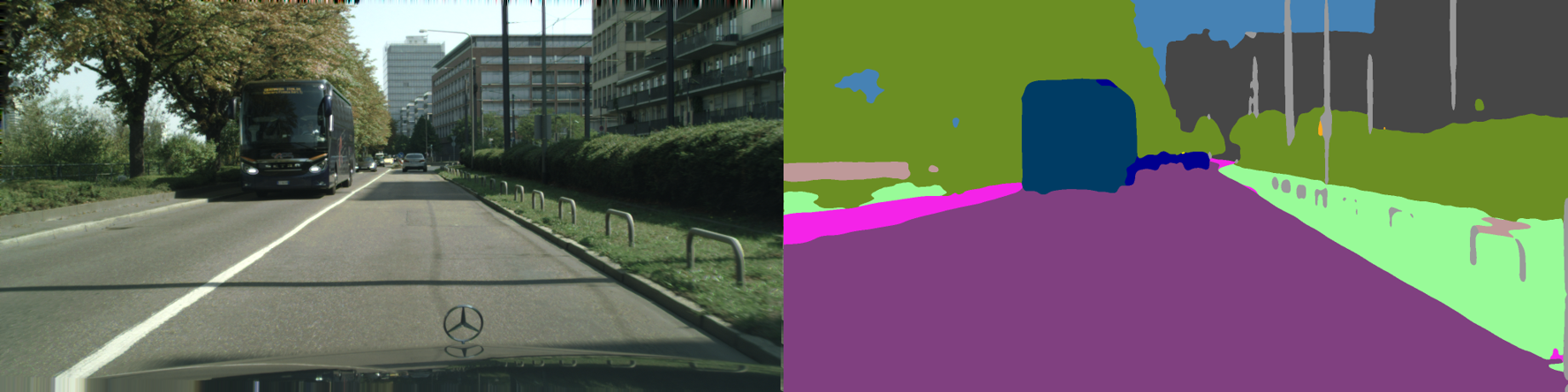}
	\caption{KittiSeg's input (left) and output (right) on the Cityscapes dataset using a Tensorflow \cite{RN327} implementation. Reprinted with permission from \cite{RN328}}
	\label{figkittiseg}
\end{figure*}

Examples of works that focuses on semantic segmentation for road scene recognition are SegNet \cite{RN272}, KittiSeg \cite{RN291} and ENet \cite{RN298}. SegNet was proposed by Badrinarayanan et al. as a CNN architecture that is often implemented on the Caffe \cite{RN273} framework. Its architecture uses an encoder-decoder network that is followed by a pixelwise classification layer, where the encoder and decoder networks consist 13 convolutional layers each. The encoder network is the same as the VGG16 \cite{RN299} network, which performs convolutions to obtain a set of feature maps. The decoder network then upsamples the feature map of each corresponding encoder, producing dense feature maps that are then batch normalised. A soft-max classifier at the output then individually classifies each pixel into one of twelve object classes to form the output image as illustrated in Fig. \ref{figsegnet}. SegNet's proposal was compared against fully convolutional network (FCN) decoding technique, also based on the VGG16 network, and DeconvNet \cite{RN300}, which uses fully connected layers. The authors tested SegNet on the CamVid dataset and SegNet's road scene segmentation results showed that while DeconvNet and SegNet yielded favourable results, SegNet's computational cost was significantly lesser due to its network being smaller. KittiSeg is the segmentation sequence of MultiNet whereby encoding is performed using the first 13 layers of the VGG16 network like SegNet. The fully connected layers of the VGG architecture is then transformed for decoding, thereby employing a fully connected network architecture. The authors used the KITTI Road Benchmark dataset for training and noted that the network converged quickly with high road segmentation efficiencies, which placed their KittiSeg on top of KITTI's road leaderboard at the time of its publication. An example of KittiSeg's input and output on the Cityscapes dataset is as illustrated in Fig. \ref{figkittiseg}. Paszke et al. noted that the VGG16 architecture that these works are based on are very large and hence less suitable for embedded and mobile applications, leading to the proposal of ENet \cite{RN298}. Since real-time semantic segmentation requires a frame rate of at least 10 frames per second (fps), this is difficult to achieve on embedded computers. ENet is a custom-designed neural network architecture proposed for high computational speed and accuracy that is designed based on ResNets \cite{RN302}.  Performance tests showed that ENet is about 17 times faster than SegNet while running on an Nvidia Jetson TX1 \cite{RN274} embedded computer, while being significantly more memory efficient. Training and testing benchmarks were performed across the Cityscapes, CamVid and SUN RGB-D \cite{RN303} datasets with their results compared against SegNet. By measuring the intersection over union (IoU) matrices, ENet was able to outperform SegNet in the Cityscapes dataset, as well as the CamVid dataset in six of its eleven classes. Treml et al. \cite{RN305} subsequently proposed a new architecture that improves on the accuracy of ENet, while being implementable on embedded computers for real-time inference. This architecture follows SegNet whereby it uses an encoder-decoder pair. The authors modified a SqueezeNet \cite{RN306} architecture for its encoder network favouring its low latencies, and a parallel dilated convolution layer \cite{RN307} as its decoder to retain high computation performance while using fewer parameters. Testing and training were performed on the Cityscapes dataset over the Caffe framework. Results were compared against ENet, outperforming it in its IoU matrices in both class and category, while compromising on slightly lower framerates on the Jetson TX1, but still exceeding the 10 fps requirement for autonomous driving. A summary of the semantic road segmentation algorithms are presented in Table \ref{tabsemantic}, listing each algorithm according to its encoder-decoder network, and the datasets used for their experiments. 

\begin{table*}[!ht]
	\renewcommand{\arraystretch}{1.3}
	\caption{Summary of semantic road segmentation algorithms presented}
	\label{tabsemantic}
	\centering
	\begin{tabular}{c c c c c}
		\hline
		\bfseries Algorithm & \bfseries Year & \bfseries Encoder & \bfseries Decoder & \bfseries Dataset\\
		\hline
		SegNet \cite{RN272}		& 2015			& VGG16  		& Custom 		& CamVid \\
		KittiSeg \cite{RN291}	& 2016			& VGG16 		& FCN 			& KITTI Road \\
		ENet \cite{RN298}		& 2016			& bottleneck 	& bottleneck 	& CamVid, Cityscapes \\
		Treml et al. \cite{RN305} &2016			& SqueezeNet 1.1 & Parallel dilated convolutions & Cityscapes \\
		\hline
	\end{tabular}
\end{table*}

\section{Commercial Implementations} \label{seccommercial}
Commercial implementations of road recognition in the automotive industry are largely based on the availability of original equipment manufacturers (OEMs) that supply advanced driver-assistance system (ADAS) computers and sensors. Mobileye \cite{RN221}, Nvidia \cite{RN174}, Velodyne \cite{RN309} and FLIR \cite{RN310} are few of the OEMs involved in manufacturing autonomous driving systems. Mobileye is reputed for their ADAS system-on-chip (SoC) called EyeQ \cite{RN311}, where in addition to lane keeping, it is capable of supporting sensor fusion, visual computing, path planning etc. towards full (Level 5 \cite{RN312}) autonomous driving while being power-efficient. Nvidia's DRIVE PX \cite{RN174} system uses their graphic processing unit (GPU) architectures to deliver on fast AI performance on mobile vehicles. They recently announced the Drive PX Xavier computer, which is an SoC that integrates a new GPU architecture, an eight-core central processing unit (CPU) and a computer vision accelerator with a 20 Watt requirement \cite{RN313}, making it ideal for real-time road recognition tasks. Nvidia has also published a work describing the mapping camera pixels to steering commands using an end-to-end approach on a CNN, which is processed by the Drive PX \cite{RN262}. Velodyne and FLIR are well-known OEMs that manufacture LiDAR and camera systems respectively for autonomous driving. Aside from OEMs, corporates that research into autonomous driving algorithms includes Google \cite{RN314} and Uber \cite{RN315}, as well as automotive manufacturers such as BMW \cite{RN316}, Volvo \cite{RN317} and Daimler \cite{RN318}.

\begin{figure}[ht]
	\centering
	\includegraphics[width=0.7\linewidth]{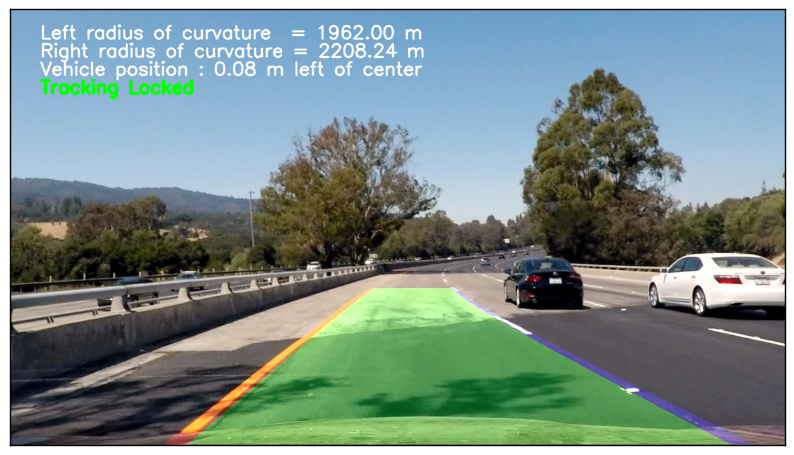}
	\caption{The Self-Driving Car Nanodegree's Term 1, Project 4 output. Reprinted with permission from \cite{RN329}.}
	\label{figudacity}
\end{figure}

Road recognition and detection techniques are also becoming more accessible to the masses. As part of an effort to produce an open-sourced autonomous car, Udacity has introduced its Self-Driving Car Nanodegree Programme, which includes road and lane detection as part of its Term 1 curriculum \cite{RN319}. Specifically, Project 1: Finding Lane Lines, and Project 4: Advanced Lane-Finding. In Project 1, students utilised OpenCV's functions such as Canny edge detection and Hough transform for road and lane detection. Project 4 expands on this to classify lane boundaries, as well as to provide the vehicle's estimated position on the road and the road's curvature. Binary images of road scenes are perspectively transformed into a birds-eye view, where lane pixels are subsequently detected for a polynomial model fitting. The model fitted lanes can then determine the road's curvature. Fig. \ref{figudacity} shows the final output of a Project 4 report, which includes marked lane boundaries with estimations of the lane curvatures and the vehicle's position.

comma.ai \cite{RN320} is a startup company by George Hotz that specialises in providing assisted and autonomous driving systems to the consumer market. Their goal is to achieve full autonomous driving with existing road vehicles with after-market devices. Most of the software that comma.ai creates are open-sourced, which includes its autonomous driving system, openpilot \cite{RN321}. openpilot performs adaptive cruise control and lane keeping that can be retrofitted to existing cars. comma.ai also includes semantic segmentation for road scenes \cite{RN322}, where they have experimented with SegNet and ENet, and proposed a solution based on ENet and ReSeg with VGG convolutional layers \cite{RN323} dubbed Suggestions Network (SugNet) to automatically label ground truths. This is in addition to the recurrent neural network approach that they took with autoencoders to learn a driving simulator as part of their initial research, which generates realistic road image predictions \cite{RN324}. 

\section{Recent Works} \label{secrecent}
Recent research developments in road detection are inclining towards supervised learning and neural networks. For instance, Brust et al. \cite{RN169} described an approach that uses an image patch that is fed into a CNN for label estimation, dubbed the Convolutional Patch Network. The image patch is used as the spatial prior for this method, which corresponds to a position of an object from a small group of pixels in the image frame. Among the other implementation methods proposed by the authors is a normalised initialisation approach to neural network parameters, thereby circumventing the vanishing gradient problem. For benchmarking, the authors used the KITTI dataset in a birds-eye view. This means that a transformation was done to convert dashboard view (ego view) images into birds-eye view, to which \cite{RN218} claimed that road detection is more efficient this way. Training weights are therefore chosen according to the pixel sizes after this conversion, as the authors noted that any classification errors that happens near the horizon pixels in ego view will escalate to many more pixels in birds eye view. Experiments performed by the authors for road detection yields a 10\% improvement over that of Alvarez et al., which is largely contributed to the addition of spatial priors into the network. Visual road recognition implementations for autonomous driving are sometimes performed on mobile robots due to local legislations and safety concerns on autonomous vehicles \cite{RN334}. \"{O}fj\"{a}ll et al. \cite{RN334} developed a road-following system that incorporates supervised and self-reinforcement learning called symbiotic online learning of associations and regression (SOLAR). The system is initially trained by a human driver with a recording camera for the system to learn the road's appearance, where upon sufficient training, the system will be capable of taking over controls from the driver. The system predicts visual feature vectors of subsequent frames using a Hebbian associate learning procedure \cite{RN335}, allowing the system to perform self-feedbacks for reinforcement learning. The authors implemented SOLAR on a remote controllable robot car in environments that simulate real roads, and subsequently compared its autonomous driving capability and learning time to qHebb \cite{RN335}, along with a CNN approach with the Caffe framework. Results showed that the CNN approach is incapable of running in real time with long learning times. Comparisons with qHebb notes that SOLAR is able to simultaneously improve learning speeds due to its reinforcement learning.  Another work that implements road detection on a mobile robot for testing is the visual road following approach by Krajn{\'i}k et al. \cite{RN219}. It shares some similarities with Crist\'{o}foris et al.'s \cite{RN156} work whereby it produces a path guide for the autonomous navigation of a mobile robot. This work, however, emphasises on their photometric methods that adjust to the reflectance of captured objects for shadow removal, hence providing an illumination invariant solution for visual road recognition. For testing, the authors implemented two threads in parallel; one for manoeuvring the robot, and the other calculates the robot's orientation with respect to the path boundary. This effectively ensures that the robot navigates at the centre of the path. An image captured by the robot's on-board camera is processed into an intrinsic image --- a process that decomposes an image into multiple layers of intrinsic properties. Using intrinsic images enables the algorithm to be illumination invariant. The authors subsequently used the intrinsic images to compute the robot's path through histogram equalisation, which segments the path from the background, thus binary classifying them into path and non-path regions. The robot's orientation for navigation is then calculated from a probability distribution estimation of intrinsic pixels from a histogram based on Shannon entropy \cite{RN220}. Experiments were conducted offline using datasets and online on a mobile robot, and the results proved that using intrinsic images allows the robot to move autonomously across different illumination conditions.

\section{Conclusion} \label{secconclusion}
This paper presented a review of literature that covers the visual road recognition process according to its associated methods, followed by methods that incorporate machine learning, and a brief review of current commercial implementations. In a typical chronology, methodological approach starts with horizon detection under the assumption that all road regions are below the horizon, which effectively isolates and segregates the sky portion above the horizon from any image processing. Subsequently, detecting vanishing points allows us to find the point of convergence between the road and the horizon, enabling horizon and road segmentation to be performed corresponding to that vanishing point. The road region below the vanishing point can be further segmented for computation efficiency by the introduction of the region of interests, which usually encapsulates the road or its edges where computation can be concatenated. Computational processes for visual road detection generally involves binary image classification, which classifies roads from non-road regions using techniques ranging from Gaussian models to histograms. Recently, many works incorporate CNN for visual classification with improved accuracies. With image classification complete, model fitting is applied to visually distinguish roads from non-road areas. There is also a recent shift in road recognition approaches using deep learning whereby semantic segmentation is increasingly utilised for road detection, along with other objects in road scenes. From a hardware perspective, visual road recognition is quickly replacing conventional methods such as using LiDAR and radar due to the rapid improvements in cost and availability of image sensors. With the general availability of datasets and libraries such as KITTI and OpenCV, along with open-source deep learning libraries such as Caffe and Tensorflow, visual road recognition is now easily implementable and evaluated even in embedded systems. Additionally, recent approaches in visual road recognition is steadily addressing the research challenges presented in this area, which encompass those that are generally found in visual computing, such as illumination invariance and camera distortions. Therefore, a robust visual road recognition system should provide high accuracies while maintaining real-time computation capabilities that is able too compensate for the dynamic changes in road scenes at any time.

\section*{Acknowledgment}

The authors would like to thank Mr Andrea Palazzi, Mr Thomas Anthony, Mr Touqeer Ahmad and Prof Hui Kong for the granting of permission to use their figures in this paper.

\bibliographystyle{unsrt}  
\bibliography{main}  

%
%
%
%

\end{document}